# Semantic Video Segmentation : Exploring Inference Efficiency

Subarna Tripathi[*], Serge Belongie[!], Youngbae Hwang[#], Truong Nguyen[*]

[*] University of California San Diego, [!] Cornell NYC Tech, [#] Korea Electronics Technology Institute

*Abstract*—We explore the efficiency of the CRF inference beyond image level semantic segmentation and perform joint inference in video frames. The key idea is to combine best of two worlds: semantic co-labeling and more expressive models. Our formulation enables us to perform inference over ten thousand images within seconds and makes the system amenable to perform video semantic segmentation most effectively. On CamVid dataset, with TextonBoost unaries, our proposed method achieves up to 8% improvement in accuracy over individual semantic image segmentation without additional time overhead. The source code is available at https://github.com/subtri/video_inference

## I. INTRODUCTION

Deep convolutional neural networks (DCNNs) trained on a large number of images with pixel-level annotations or a combination of strongly labeled and weakly-labeled images have recently been the state-of-the-art in semantic image segmentation, with significant performance improvement. However, due to the very invariance properties that make DCNNs good for high level tasks such as classification, visual delineation capacities for deep learning techniques are limited. Recent approaches address this problem with Conditional Random Field (CRF) based graphical model [5] in two ways: either by adding a post-processing step [3, 8] of CRF-based probabilistic graphical model for the pixel-level classification or, by integrating the graphical model as a part of the CNN to enable end-to-end learning [11] with the usual back-propagation possible without the need of post-processing.

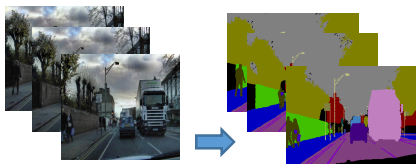

*Figure 1: Video Semantic Segmentation problem*

Alvarez *et al.* [1] demonstrates that performing inference on all test images at once in a dense CRF yields better results than inferring one image at a time without additional computation cost compared to performing segmentation sequentially on individual images. Yet, it lacks the ability to handle high-order terms such as label consistency over large regions (pattern-based potentials) and relations of global co-occurrence potentials, are shown to be more expressive and effective for object class segmentation task [4, 6, 7]. Filter-based inference for those higher-order terms is formulated in [10] which enables significant speed-up. Yet, it needs to consider temporal consistency when applied in co-segmentation or video semantic segmentation. We explore the efficiency of the CRF inference module beyond image level semantic segmentation (Figure 1). The key idea is to combine semantic co-labeling and exploiting more expressive models. Similar to [1], our formulation enables us to perform inference over ten thousand images within seconds. On the other hand, it can handle higher-order clique potentials similar to [10] in terms of region-level label consistency and context in terms of co-occurrences. We follow the mean-field updates for higher order potentials and extend the spatial smoothness and appearance kernels to address co-segmentation of multiple frames of a video; thus making the system amenable to perform video semantic segmentation most effectively.

## II. METHODOLOGY

### A. Joint labelling on video frames

In this work, we deal with multiple images $\mathbf{I}_1, ..., \mathbf{I}_M$ and the random field is defined over the set $\mathbf{Y} = \{\mathbf{X}^1, ..., \mathbf{X}^M\}$, where superscript denotes index in image sequence. MAP labeling can be found by computing the assignment that maximizes the distribution or equivalently minimizes the energy function, $E$

$$E(\{X^1, ..., X^F\}|I^1, ..., I^F) = \sum_{\substack{i \in \mathcal{F} \\ j \in \aleph}} \psi_u(x_j^i) + \sum_{\substack{i,i' \in \mathcal{F} \\ j,j' \in \aleph}} \psi_p(x_j^i, x_{j'}^{i'}) \quad (1)$$

where $\mathcal{F} = \{1, ..., F\}$ and $\aleph = \{1, ..., N\}$, and $\psi_u(.)$ and $\psi_p(.,.)$ denote the unary and pairwise potentials respectively. The unary potential function encodes the cost of assigning a specific label to a pixel.

At every iteration of the algorithm, the update for each variable involves summing over all the random variables. However, with a Gaussian kernel, the expensive summation becomes equivalent to performing high-dimensional Gaussian filtering [5] and thus remains completely tractable. Thus the pairwise potential functions take the form:

$$\psi_p(x_j^i, x_{j'}^{i'}) = \mu(x_j^i, x_{j'}^{i'}) \sum_{m=1}^M w_m \, k_m(v_j^i, v_{j'}^{i'}) \quad (2)$$

where $\mu(.,.)$ is a label compatibility function. This encodes a Potts model, that is $\mu(x_j^i, x_{j'}^{i'}) = \mathbf{1}[x_j^i \neq x_{j'}^{i'}]$. The kernel $k_m(.,.)$ is a Gaussian Kernel computed over feature vector $v_j^i$ that describes pixel j in image *i*. We utilize spatial smoothness and appearance kernels $k_1$ and $k_2$ as:

$$k_1(v_j^i, v_{j'}^{i'}) = \exp\left(-\frac{\|p_j^i - p_{j'}^{i'}\|^2}{\sigma_p^2} - \frac{\|i - i'\|^2}{\sigma_f^2}\right) \quad (3)$$

$$k_2(v_j^i, v_{j'}^{i'}) = \exp\left(-\frac{\|p_j^i - p_{j'}^{i'}\|^2}{\sigma_l^2} - \frac{\|i - i'\|^2}{\sigma_t^2} - \frac{\|I_j^i - I_{j'}^{i'}\|^2}{\sigma_c^2}\right) \quad (4)$$

where $p_j^i$ and $I_j^i$ encode the image location and color vector of pixel j in image i. Setting $\sigma_f$ to a low value will make the kernel $k_1$ vanish for any two random variables not belonging to the same image. Mean-field updates are calculated using ConCaveConvexProcedure (CCCP) which is explained in [5].

This work is supported by the Technology Development Program for Commercializing System Semiconductor funded by the Ministry Of Trade, Industry and Energy (MOTIE, Korea). (No 10041126, Title: International Collaborative R&BD Project for System Semiconductor)

Like [5], our approach approximates this operation using the permutohedral formulation. This performs convolution on a down-sampled version of the graph. Thus the overall cost becomes linear with the input size. The cost of performing inference in the CRF through our approach which exploits all images simultaneously is the same as the cost of performing inference sequentially on the individual images.

*B. Higher-order Clique Potentials*

For high-order clique's potentials, the general energy takes the form as shown below:

$$E(V|I) = \sum_{c \in C} \psi_c(v_c|I) \quad (5)$$

In general, X ⊆ V when higher-order cliques (HOC) are taken into consideration. The most useful higher-order terms for object detection task are pattern-based potential and co-occurrence potentials. For still images, patterns based potentials are coming from the different number of superpixels. We consider the use of patterns from either superpixels or the slices of supervoxels corresponding to different video frames, thus enforcing temporal consistency. Mean-field update for these high-order terms follows the formulation described in [10]. Thus, inference in video can exploit intra and inter-frame connectivity and also respect patterns based potentials which are time consistent.

## III. RESULTS

We evaluate our approach on standard benchmark dataset for multi-class segmentation: the Cambridge-driving Labeled Video dataset (CamVid) [2]. CamVid consists of four image sequences with ground truth labels at 1fps that associate each pixel with one of 32 semantic classes. All the experiments were conducted using single threaded code on a standard Intel(R) Core(TM) i7-3770 CPU 3.40GHz desktop. In our current experiments, we used the TextonBoost [9] unary potentials for easy comparison with other recent methods. For pattern-based potentials, we use three different superpixel segmentations by varying parameters of the meanshift algorithm. Frame-level Dense-CRF with this $P^n$-Potts model [10] almost achieves similar quality as of previous graph-cut based slow inference method [6], but lacks temporal consistency.

*Table 1. Average per-class accuracy on CamVid dataset*

| [5] | [1] | [10] | ours |
|-----|-----|------|------|
| 47.9 | 48.8 | 55.3 | 56.8 |

The proposed video-level Dense-CRF with $P^n$-Potts model shows improved temporal consistency over the frame-level operation (previous row) without additional time-overhead. Video-Level dense CRF [1] and the proposed method perform inference on 50 frames at once. On CamVid, with TextonBoost unaries, our proposed method achieves 8% more accuracy than [1] by virtue of $P^n$-Potts model and 1.5% more accuracy over [10] without additional time overhead by virtue of co-labeling (Table 1). Average per-class accuracy is computed as the average over all classes of the ratio of correctly classified pixels in a class to the total number of pixels in that class.

## IV. CONCLUSIONS

In this work, we explore the idea of combing the best of co-labeling and exploiting more expressive models for efficient inference in Conditional Random Field. This framework achieves state-of-the art semantic segmentation performance in CamVid dataset and opens up possibility of improving performance of deep learning based video semantic segmentation with this inference engine. CNN feature classification yields better unary potentials compared to the unaries provided by TextonBoost as evidenced by recent works [3, 5, 8]. Analyzing the final video semantic segmentation accuracy using CNN based unaries and proposed dense-CRF with $P^n$-Potts model remains our future work. The source code is available at https://github.com/subtri/video_inference.